\documentclass{article}

\usepackage [preprint] {neurips_2023}

\usepackage[utf8]{inputenc} 
\usepackage[T1]{fontenc}    
\usepackage{hyperref}       
\usepackage{url}            
\usepackage{booktabs}       
\usepackage{amsfonts}       
\usepackage{nicefrac}       
\usepackage{microtype}      
\usepackage[table]{xcolor}         
\usepackage{amsmath,amssymb,amsthm}
\usepackage{graphicx}
\usepackage[normalem]{ulem}
\usepackage{bbm}
\usepackage{comment}
\usepackage{multirow}
\usepackage{caption}
\usepackage{subcaption}
\usepackage{siunitx}
\usepackage{etoolbox}
\usepackage{algorithm}
\usepackage{algorithmic}

\usepackage[noabbrev,nameinlink,capitalise]{cleveref}

\theoremstyle{plain}
\newtheorem{theorem}{Theorem}[section]
\newtheorem{proposition}[theorem]{Proposition}
\newtheorem{lemma}[theorem]{Lemma}
\newtheorem{corollary}[theorem]{Corollary}
\theoremstyle{definition}
\newtheorem{definition}[theorem]{Definition}

\theoremstyle{remark}

\title{Online Policy Learning from Offline Preferences}

\author{%
  Guoxi Zhang\thanks{This work was done when the author was a student at Kyoto University.} \\
  Kyoto University\\
  Kyoto, Japan  \\
  \texttt{guoxi@ml.ist.i.kyoto-u.ac.jp} \\
  \And
  Han Bao \\
  Kyoto University\\
  Kyoto, Japan  \\
  \texttt{bao@i.kyoto-u.ac.jp} \\
  \And 
  Hisashi Kashima \\
  Kyoto University\\
  Kyoto, Japan  \\
  \texttt{kashima@i.kyoto-u.ac.jp} \\
}

\begin{document}

\maketitle

\begin{abstract}
In preference-based reinforcement learning (PbRL), a reward function is learned from a type of human feedback called preference.
To expedite preference collection, recent works have leveraged \emph{offline preferences}, which are preferences collected for some offline data.
In this scenario, the learned reward function is fitted on the offline data.
If a learning agent exhibits behaviors that do not overlap with the offline data, the learned reward function may encounter generalizability issues. 
To address this problem, the present study introduces a framework that consolidates offline preferences and \emph{virtual preferences} for PbRL, which are comparisons between the agent's behaviors and the offline data. Critically, the reward function can track the agent's behaviors using the virtual preferences, thereby offering well-aligned guidance to the agent.
Through experiments on continuous control tasks, this study demonstrates the effectiveness of incorporating the virtual preferences in PbRL.
\end{abstract}

\section{Introduction}
Preference-based reinforcement learning (PbRL) is a setting for developing agents using human preferences~\citep{10.1007/978-3-642-23780-5_11}.
A preference can be an outcome of comparisons between a pair of actions~\citep{10.1007/s10994-012-5313-8}, states~\citep{6861960}, or trajectories~\citep{NIPS2017_d5e2c0ad}, for the extent they meet task specifications.
PbRL is intriguing for two primary reasons.
First, as pointed out by~\citet{thrustone}, pairwise comparisons are less subjective than absolute scoring, allowing preferences to be collected from people who cannot quantitatively evaluate agents.
This is a desirable property for human-involved tasks such as value alignment~\citep{10.1007/978-3-030-28619-4_7} and shared autonomy~\citep{DBLP:conf/rss/ReddyDL18}.
Furthermore, preference collection is seamlessly scalable, as a typical preference query requires comparing videos for only a few seconds~\citep{NIPS2017_d5e2c0ad}, which allows for collecting a large amount of preferences economically using crowdsourcing.

\begin{figure}[t]
  \centering
  \newdimen\figrasterwd
  \figrasterwd\textwidth
  \parbox{\figrasterwd}{
    \parbox{.49\figrasterwd}{%
      \subcaptionbox{A diagram for online PbRL. Annotators must wait for behaviors to be generated during policy learning. \label{fig:online_pbrl} }{\includegraphics[width=\hsize]{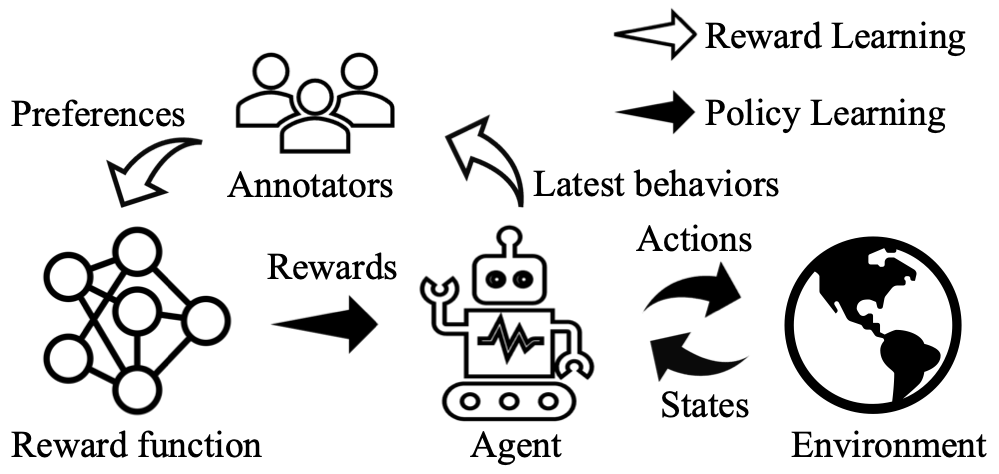}}
      \subcaptionbox{A diagram for using offline preferences. All preferences are collected before policy learning.\label{fig:offline_pbrl}}{\includegraphics[width=\hsize]{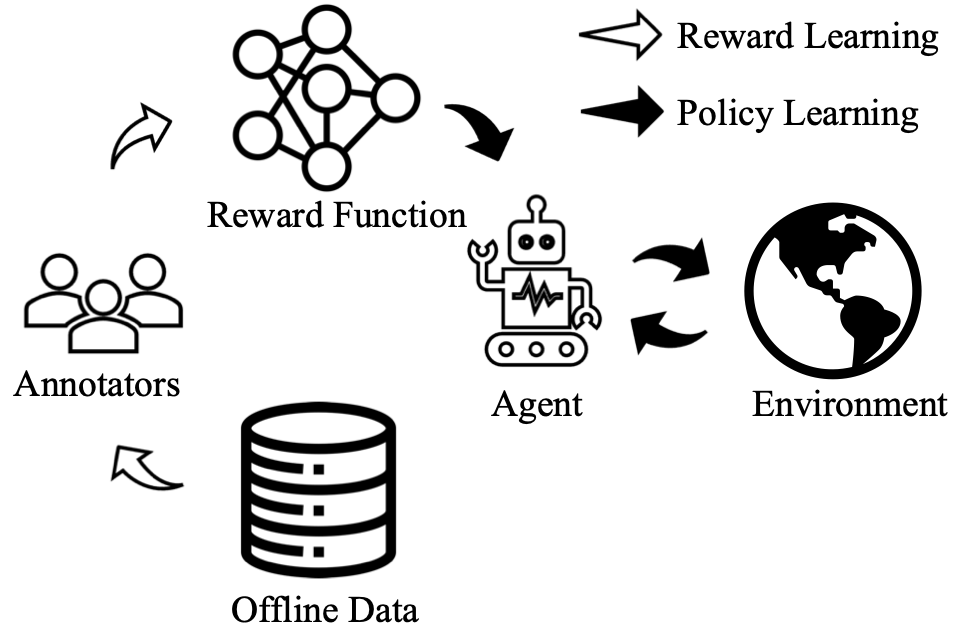}}  
    }
    \hskip1em
    \parbox{.49\figrasterwd}{%
      \subcaptionbox{\textbf{Upper left}: An example for behaviors in the offline data.
      The robot arm is pushing the white cylinder to the red spot.
      \textbf{Middle right}: An example for the agent's behaviors in the initial phase of policy learning.
      The robot arm cannot is far away from the white cylinder.
      There is little overlap between the state distribution of the offline data and that of the agent's initial behaviors.
      So reward functions learned from offline preferences may not align with the agent's performance.\label{fig:behavior_dist}}{\includegraphics[width=\hsize]{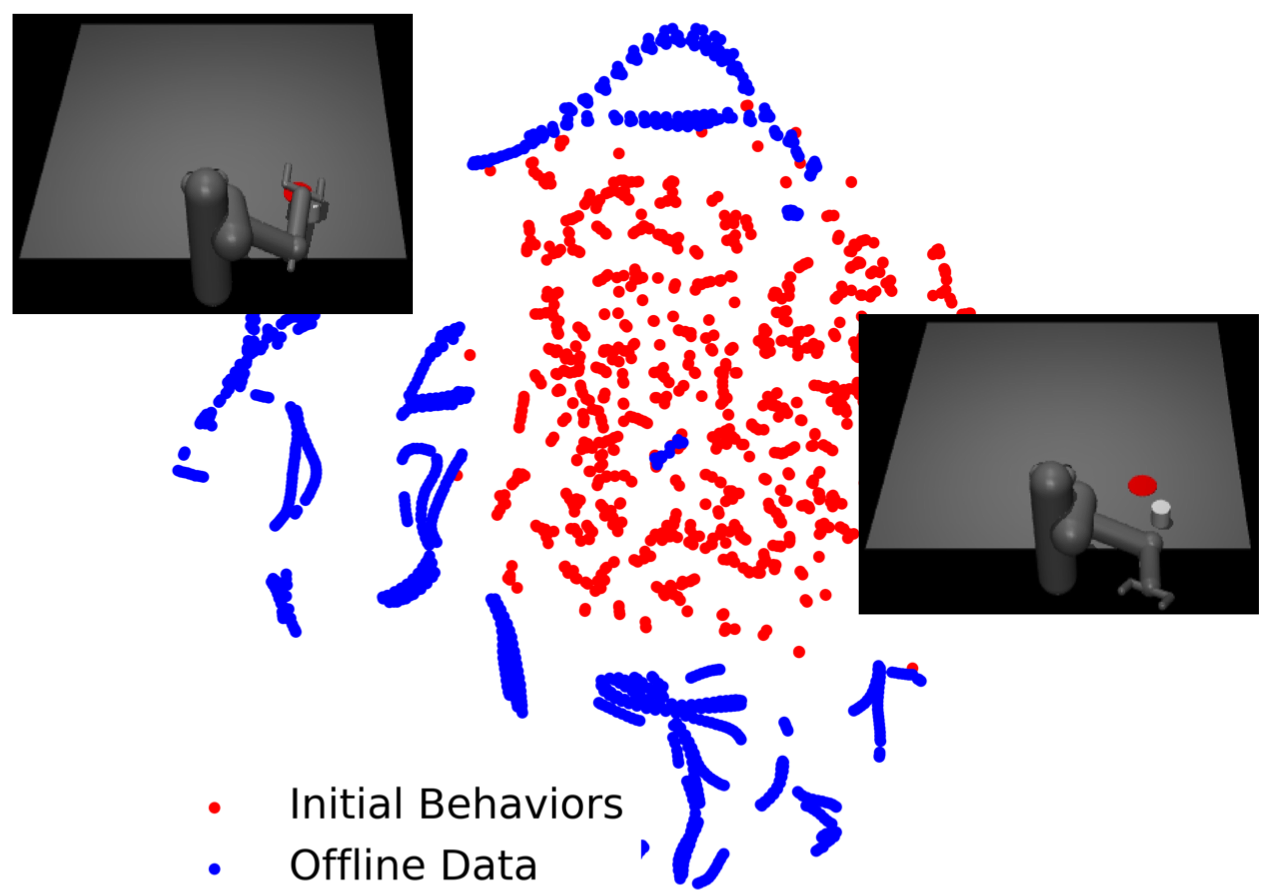}}
    }
    \caption{A diagram for online PbRL (\textbf{(a)}), learning from offline preferences (\textbf{(b)}), and the generalizability problem of using offline preferences (\textbf{(c)}).}
  }
\end{figure}

In its online formulation (\cref{fig:online_pbrl}), PbRL requires on-demand assessments from humans.
This is inefficient in terms of human time, because annotators are entirely occupied during policy learning, mostly waiting for trajectories.
A recent idea is to adopt \emph{offline preferences} for better efficiency~\citep{daniel-2021}, which means to collect preferences for certain existing trajectories (called offline data), as illustrated in \cref{fig:offline_pbrl}.
However, one caveat is that, there may be a distribution shift between the offline data and the agent's behaviors.
For example, consider the Pusher task from the gym library~\citep{1606.01540}, where the agent controls a robot arm to push a white cylinder to a red spot.
The states in offline data (\cref{fig:behavior_dist} upper left) are clustered, likely due to the limited number of feasible configurations for the robot's joints.
Meanwhile, the states of the agent's initial behaviors (\cref{fig:behavior_dist} middle right) are disributed uniformly. 
When the distribution shift exists, we may face a \emph{generalizability problem}. 
As the agent's reward function is only trained for behaviors in the offline data, it might not generalize to the agent's behaviors.
In our experiments for Pusher, the learned reward function cannot predict the ranking of the agent's behaviors (``PbRL'' in \cref{fig:rank_correlation}), leading to poor task performance (``PbRL'' in \cref{tab:results_mixture_all}).
Since we cannot control the distribution of the offline data in practice, this generalizability problem risks practical use of offline preferences.

To address the generalizability problem, we propose a framework called preference-based adversarial imitation learning (PbAIL).
The key idea is to generate \emph{virtual preferences} that favor offline data over the agent's behaviors. 
By jointly maximizing the likelihood of offline and virtual preferences, PbAIL can learn a reward function that aligns with the agent's behaviors.
In the meantime, it also improves the agent's policy using the learned reward function.
The reward learning step and the policy learning step are alternated to ensure the reward function always align with the agent's behaviors, which is equivalent to solving a max-min objective.
Since offline data can be imperfect in practice, we extend PbAIL to model the reliability of virtual preferences to handle imperfect data.

This work evaluated PbAIL from three perspectives: (i) its task performance on imperfect offline data and preferences, (ii) the individual effect of learning from virtual preferences and modeling their reliability, and (iii) how its performance changes with preference size and offline data quality.
In our experiments for seven Mujoco tasks, PbAIL consistently achieves good performance when compared to existing approaches that use both offline data and preferences~\citep{10.5555/3327757.3327897,NEURIPS2021_670e8a43}.
In particular, PbAIL can achieve better performance when compared to only using offline preferences in six of the seven tasks.
In an ablation study, we confirm that it matches the return of offline data when using virtual preferences, and it achieves better performance on imperfect offline data when modeling the reliability of virtual preferences. 
Lastly, our results for the effect of preference size and data quality highlight that PbAIL is suitable when preferences are more accessible than high-fidelity trajectories.
In summary, our contributions are as follows:
\begin{itemize}
    \item We propose PbAIL to overcome the generalizability problem that arises when learning reward functions from offline preferences.
    \item We extend PbAIL to handle imperfect offline data.
    \item We extensively evaluate PbAIL for non-optimal offline data and limited preferences, clarifying its strength, the effects of its components, and its limitations. 
\end{itemize}
\noindent The rest of this paper is organized as follows.
\Cref{sec:related_work} reviews related literature, while \cref{sec:preliminaries} provides background knowledge. \Cref{sec:proposed_framework} formulates the learning problem and introduces the proposed PbAIL. \Cref{sec:experiments} presents empirical results, and \cref{sec:conclusion} concludes the paper.

\section{Related Work}\label{sec:related_work}
PbRL has been studied for over a decade~\citep{10.1007/978-3-642-23780-5_11,10.1007/s10994-012-5313-8} and applied to Atari games~\citep{NIPS2017_d5e2c0ad}, locomotion tasks~\citep{lee2021bpref}, navigation tasks~\citep{daniel-2021}, and fine-tuning language models~\citep{ouyang2022training}.
Recent advancements include enhancing exploration~\citep{pmlr-v139-lee21i}, adaptive query selection~\citep{DBLP:conf/iros/WildeK020,biyik2020active}, and improving feedback efficiency~\citep{park2022surf,liu2022metarewardnet}.
Specifically, utilizing offline preferences \citep{daniel-2021,crowd_pbrl} allows for more efficient use of annotator's time.
What remains a question is its impact on reward learning, especially if there is a mismatch between the distributions of offline data and agents' behaviors.

In PbRL literature, the approach proposed by~\citep{10.5555/3327757.3327897} is related to this work, as it combines PbRL and behavior cloning (BC).
However, this approach cannot handle imperfect offline data.
Meanwhile, the confidence-aware imitation learning (CAIL) algorithm~\citep{NEURIPS2021_670e8a43} accepts both offline data and preferences as input, but it does not maximize the likelihood of preferences.
We empirically compare PbAIL with these two approaches in \cref{sec:experiments}.
\section{Preliminaries}\label{sec:preliminaries}

\subsection{Reinforcement Learning}
\paragraph{MDP}
Reinforcement learning (RL) uses the Markov decision process $\langle\mathcal{S}, \mathcal{A}, P, r, \gamma, \mu \rangle$~\citep{10.5555/3312046} to model sequential decision-making tasks. 
Here, $\mathcal{S}$ is the set of states, and $\mathcal{A}$ is the set of actions; they represent information and options available to an agent for decision-making, respectively.
The transition probability $P:\mathcal{S}\times\mathcal{A}\to\mathit{\Delta}(\mathcal{S)}$ governs how states transit, where $\mathit{\Delta}(\cdot)$ is the set of distributions over a set.
The reward function $r:\mathcal{S}\times\mathcal{A}\rightarrow \mathbb{R}$ is a function that evaluates agents' decisions, which is unknown and to be inferred in PbRL.
The discount factor $\gamma\in (0,1)$ is later used to define the value function, and $\mu$ is the distribution for initial states.
An MDP prescribes a protocol for sequential interaction between an agent and a hypothetical entity called the environment.
Starting from an initial state sampled from $\mu$, the agent observes a state $s\in\mathcal{S}$ from the environment and selects an action $a$ according to a stochastic policy $\pi:\mathcal{S}\to \mathit{\Delta}(\mathcal{A})$ associated with the agent.
It then receives the next state from the environment, which is sampled from $P(\cdot|s,a)$.
A sequence of states and actions generated in interaction $(s_1, a_1, s_2, a_2, \dots)\overset{\mathrm{def}}{=}\tau$ is defined as a trajectory.
To simplify notations, this work occasionally uses $x$ as a shorthand notation for a state-action pair $(s,a)$.

\paragraph{RL Objective}
The \emph{return} of a trajectory is the sum of rewards assigned to its state-action pairs.
For state $s\in\mathcal{S}$, the \emph{value function} $v^\pi(s)$ of a policy $\pi$ specifies the expectation of $\gamma$-discounted return starting from an initial state $s$ and following a policy $\pi$.
It is defined as follows:
\begin{equation*}
    v^\pi(s) \overset{\mathrm{def}}{=} \mathbb{E} \left[\sum_{t=1}^\infty \gamma^{t-1}r(s_t,a_t) \,\middle|\, \pi, s_1=s\right],
\end{equation*}
where the expectation is taken over states (following the transition probability) and actions (following the policy).
The \emph{value} of a policy $\pi$ is the expectation of $v^\pi(s)$ over initial-state distribution $\mu$: $v^{\pi}=\mathbb{E}_{s\sim \mu}[v^\pi(s)]$. 
The goal of RL is to find an optimal policy $\pi^*$ such that $v^{\pi^*}\geq v^\pi$ for any policy $\pi$ under a given reward function $r$.

A useful quantity is the discounted occupancy measure, which is defined as follows.
\begin{definition}[\citet{10.5555/528623}]
The \emph{discounted occupancy measure} $\rho^\pi$ for policy $\pi$ is defined as
\begin{equation*}
    \rho^{\pi}(s,a)=\sum_{t=1}^\infty \gamma^{t-1}\Pr(s_t=s,a_t=a;\pi),
\end{equation*}
where $\Pr(s_t=s,a_t=a;\pi)$ is the probability of the joint event $s_t=s$ and $a_t=a$ when following the transition probability and policy $\pi$.
\end{definition}
The discounted occupancy measure $\rho^{\pi}$ can be interpreted as an unnormalized measure for state-action pairs generated by $\pi$.
It is straightforward to show that the normalizer of occupancy measures is $\frac{1}{1-\gamma}$.
\begin{corollary}
$\sum_x \rho^\pi(x) = \frac{1}{1-\gamma}$ for any policy $\pi$.
\end{corollary}
For function $f: \mathcal{S}\times\mathcal{A} \to \mathbb{R}$, we write $\mathbb{E}_{x \sim \rho^{\pi}}[f(x)]$ as the sum of $f(x)$ over all state-action pairs weighted by $\rho^\pi(x)$ with slight abuse of notation.
The values can be expressed using $\rho^\pi$ alternatively.
\begin{corollary}
$v^{\pi}=\mathbb{E}_{x\sim \rho^{\pi}}[r(x)]$.
\label{eq:occ_value}
\end{corollary}

As a final remark, the following theorem relates policies with occupancy measures, which allows one to regard policies as samplers for state-action pairs. 
\begin{theorem}[\citet{10.1145/1390156.1390286}] Suppose $\rho$ is an occupancy measure and $\pi\overset{\text{def}}{=}\frac{\rho(s,a)}{\sum_a \rho(s,a)}$.
Then, $\rho$ is the occupancy measure for $\pi$, and $\pi$ is the only policy whose occupancy measure is $\rho$.
\end{theorem}

\subsection{Preference-based Reinforcement Learning}
\label{sec:pbrl}

\paragraph{Thurstone's model~\citep{thrustone}}
Preferences are formally defined using \emph{Thurstone's model}~\citep{thrustone}.
Suppose there is a general object space $\mathcal{O}$.
For each object $o\in\mathcal{O}$, this model assumes a ``true'' utility score $G(o)\in\mathbb{R}$. 
Annotators compare objects using disturbed utility $G(o)+\epsilon$, where $\epsilon$ is sampled from some distribution for noise.
Then, the probability of object $o$ being preferred over object $o'$ follows $\Pr(o \succ o') = \Pr(G(o)-G(o') > \epsilon'-\epsilon)$, where $\epsilon'$ is the noise associated with $o'$.
Subsequently, we work with a special case of Thurstone's model called \emph{Bradley--Terry model} (BT model)~\citep{10.2307/2334029}.
Assuming $\epsilon \sim \mathrm{Gumbel}(0,1)$, we have $\Pr(o \succ o') = \sigma(z - z')$, where $\sigma(\cdot)$ is the sigmoid function.
Under this model, the log-likelihood of a preference $o\succ o'$ is denoted by $\ell(o,o')=\log\sigma(G(o)-G(o'))$.

\paragraph{Preference-based Reward Learning}

We assume that the offline preferences are specified between trajectories, as they provide annotators with more information than states or state-action pairs.
Under the BT model, a trajectory $\tau$ is preferred over another trajectory $\tau'$, denoted as $\tau\succ\tau'$, if the utility of $\tau$ is higher than that of $\tau'$.
This means that $\tau$ corresponds to a more desirable outcome from an annotator's perspective.
We will use preferences between state-action pairs to formulate the proposed virtual preferences.
In this case, a state action pair $x$ is preferred over another state-action pair $x'$ if the utility of $r(x)>r(x')$.

Reward functions can be learned by maximizing the likelihood of preferences under the BT model.
A canonical assumption for parameterizing trajectory utility is the sum of rewards in a trajectory: $\sum_{x\in\tau}r(x)$~\citep{NIPS2017_d5e2c0ad,10.5555/3327757.3327897}.
We use the reward of a state-action pair as its utility.
A sample for trajectory preferences is written as $(\tau_{1}, \tau_{2}, c)$, where $\tau_{1}$ and $\tau_{2}$ are the trajectories being compared.
$c=1$ if $\tau_{1} \succ \tau_{2}$, and $c_j=0$ otherwise. 
Suppose the reward function is parameterized by $\theta_r$, and let the sum of estimated rewards for state-action pairs in a trajectory $\tau$ to be $G(\tau;\theta_r)$.
Given a set of $M$ preferences $\mathcal{Y}$, we can learn $\theta_r$ by minimizing:

\begin{equation}
        L_{\mathrm{pref}}(\theta_r) =  -\frac{1}{M}\sum_{(\tau_1, \tau_2, c)\in\mathcal{Y}} \left[c \log(\sigma(G(\tau_1;\theta_r)  
        -G(\tau_2;\theta_r)))+ (1-c) \sigma(G(\tau_2;\theta_r)  
        -G(\tau_1;\theta_r))\right].
        \label{eq:pref_obj}
\end{equation}

\section{Preference-based Adversarial Imitation Learning}\label{sec:proposed_framework}
This section presents the proposed PbAIL framework.
After formulating the learning problem, we describe how PbAIL leverages offline data for reward learning in \cref{subsec:learning_from_demo} and how it handles imperfect data in \cref{subsec:handling_imperfection}.

\subsection{Problem Setup}
An agent is provided with a set of $M$ offline trajectory preferences $\mathcal{Y}$, which are collected for $N$ trajectories $\mathcal{D}$ generated by behavior policy $b$.
To align the learned reward function with the agent's behaviors, we assume the agent has access to $\mathcal{D}$. 
The agent is supposed to learn a reward function $r$ and a policy $\pi$.
Similar to online PbRL, new trajectories can be generated during policy learning; however, no new real preferences may be collected.

\subsection{Reward Learning from Offline Trajectories}\label{subsec:learning_from_demo}
First, we borrow the notion of \emph{policy preferences}~\citep{10.1007/s10994-014-5458-8} to draw a connection between the behavior policy and reward learning.
A policy $\pi$ is preferred over another policy $\pi'$, denoted as $\pi\succ\pi'$, if the disturbed utility of $\pi$ is larger than that of $\pi'$ under the BT model.
PbAIL assumes that $b$ is preferred over any other policy, i.e., $b\succ\pi,\forall \pi$.
Assuming the utility of a policy $\pi$ is its value $v^\pi$, the log-likelihood of $b\succ\pi$ is given by $\log(\text{Pr}(b\succ\pi))=\log(\sigma(v^b-v^\pi)$).
To ensure $b\succ\pi$ holds for all $\pi$, PbAIL maximizes the worst-case log-likelihood of policy preference as follows:
\begin{equation}
    \max_r \min_\pi \log\sigma(v^b-v^\pi).
    \label{eq:value_maxmin}
\end{equation}
This objective involves learning a reward function $r$ and a policy $\pi$.
The maximization over $r$ enlarges the difference between $v^b$ and $v^\pi$ to make $b$ more preferable.
The minimization over $\pi$ is equivalent to maximizing $v^\pi$ with respect to $r$, which can be solved by RL.
This work uses soft actor-critic (SAC)~\citep{Haarnoja2018} as the policy learning algorithm, since it employs the principle of maximum entropy to enhance exploration.

While the maximization over reward functions in \cref{eq:value_maxmin} can be solved by approximating values using sampled trajectories, such a direct approach is subject to high variance.
This work thus proposes an efficient optimization based on the following tight lower bound.
By noting that $\log\sigma(\cdot)$ is concave and the value $v^{\pi}$ can be expressed as $v^{\pi} = \mathbb{E}_{x \sim \rho^{\pi}}[r(x)]$ (\cref{eq:occ_value}), the following inequality holds from Jensen’s inequality:
\begin{equation}
    \log\mathrm{Pr}(b\succ\pi) \geq \mathbb{E}_{\substack{x\sim \rho^b, x'\sim\rho^{\pi}}} \ell (x, x')\overset{\text{def}}{=}U_b(\pi, r).
  \label{eq:policy_pref_lowerbound}
\end{equation}
The likelihood lower bound $U_b(\pi, r)$ can be efficiently approximated using state-action pairs.
We can use samples in $\mathcal{D}$ to estimate the expectation over $\rho^b$.
When using off-policy RL backbones such as SAC, we can estimate the expectation over $\rho^\pi$ using state-action pairs in the replay buffer of the RL backbone. Let the replay buffer be $\mathcal{D}_\text{RL}$
Then, the reward maximization in \cref{eq:value_maxmin} is approximated by minimizing the following loss function:
\begin{equation}
        L_\text{virtual}(\theta_r)=-\mathbb{E}_{\substack{x\in \mathcal{D}\\ x'\in \mathcal{D}_\text{RL}}}\log(\sigma(r(x;\theta_r)-r(x';\theta_r))).
    \label{eq:virtual_obj}
\end{equation}

Given both offline preferences $\mathcal{Y}$ and trajectories $\mathcal{D}$, we can minimize $L_\text{pref}(\theta_r)+L_\text{virtual}(\theta_r)$ to ensure the reward function learned from offline preferences to align with the agent's behaviors, thereby overcoming the generalizability issue. 

\paragraph{Virtual Preferences}
Note that \cref{eq:virtual_obj} coincides with learning from state-action preferences $x\succ x'$ such that $x\in\mathcal{D}$ and $x'\in\mathcal{D}_\text{RL}$.
By minimizing \cref{eq:virtual_obj}, we are maximizing the log-likelihood that state-action pairs in offline data $\mathcal{D}$ are preferred over state-action pairs generated by the agent.
Since such preferences are not collected from annotators, we consider them to be generated by a \emph{virtual annotator} and call them \emph{virtual preferences}.
We use $\succ_v$ for virtual comparisons and $\mathcal{Y}_v$ for virtual preferences.
The use of virtual state-action preferences is a key idea of PbAIL.
As will be discussed in \cref{subsec:handling_imperfection}, it leads to a straightforward approach for handling non-optimal offline data.

As a final remark, it is straightforward to combine recent ideas of PbRL, such as the use of data augmentation~\citep{park2022surf} or active query generation~\citep{biyik2020active}, with PbAIL. We leave these extensions as future work and focus solely on showing the effectiveness of PbAIL.

\subsection{Handling Imperfect Data}\label{subsec:handling_imperfection}
In practice, we may encounter imperfect offline data.
In this case, adapting the reward function using offline data by minimizing \cref{eq:virtual_obj} becomes problematic. 
Based on the interpretation of virtual preferences, we propose to handle imperfect data by inferring the reliability of virtual preferences and the reward function simultaneously.
Our approach is based on a probabilistic model for noisy preferences collected from multiple annotators~\citep{crowd_pbrl}.
We simplify the model since the virtual preferences are not collected from multiple annotators.

Specifically, we assume that to generate a preference for $x\in\mathcal{D}$ and $x'\in\mathcal{D}_\text{RL}$, the virtual annotator first generates a temporary label using the BT model and the ground-truth reward function.
Then, it reports the label correctly with probability $\alpha(x,x')$ and incorrectly with probability $1-\alpha(x,x')$.
By the law of total probability, the probability for $x\succ_v x$ under this model, denoted by $\text{Pr}_\text{imperfect}(x\succ_v x')$, is given by
\begin{equation}
    \text{Pr}_\text{imperfect}(x\succ_v x') = \alpha(x,x')\sigma(r(x)-r(x')) + (1-\alpha(x,x'))\sigma(r(x')-r(x)).
    \label{eq:unreliable_virtual_prob}
\end{equation}
\noindent Suppose $\alpha$ is parameterized with neural network $\theta_\alpha$. Under this model, objective for learning from virtual preferences becomes
\begin{equation}
    \begin{split}
        L_{\substack{\text{imperfect}\\\text{virtual}}}(\theta_r,\theta_\alpha)=-&\mathbb{E}_{\substack{x\in\mathcal{D} \\ x'\in\mathcal{D}_\text{RL}}} \log (\text{Pr}_\text{imperfect}(x\succ_v x')).
    \end{split}
    \label{eq:unreliable_virtual_obj}
\end{equation}

Essentially, we model the reliability of virtual preferences using $\alpha(x,x')$. 
We use the reward difference $r(x)-r(x')$ and the index for the trajectory from which $x'$, denoted as $I(x')$, as features for this reliability network. 
$r(x)-r(x')$ is informative as it suggests the temporary label in the generative process of this probabilistic model. 
As for $I(x')$, recall that we sample state-action pairs from the replay buffer $\mathcal{D}_\text{RL}$ for better sample efficiency.
This buffer stores state-action pairs generated in the entire policy learning process.
Since the policy is generally improving during this process, state-action pairs generated during late stage of policy learning are more likely to be better than state-action pairs in $\mathcal{D}$, when compared to state-action pairs generated earlier.


\paragraph{Initialization} As suggested by ~\citet{crowd_pbrl}, this model needs an initialization phase.
The intuition is that, since the agent's policy is initialized from scratch, in the initial phase of policy learning the offline data are likely to be better than the agent's behaviors.  
The objective for this phase is \cref{eq:unreliable_virtual_obj}:
\begin{equation}
    \begin{split}
        L_{\substack{\text{imperfect}\\\text{virtual,init}}}(\theta_r,\theta_\alpha)=L_\text{virtual}(\theta_r)-&\mathbb{E}_{\substack{x\in\mathcal{D} \\ x'\sim\rho^{\pi}}} \log (\alpha(x,x')).
    \end{split}
    \label{eq:unreliable_virtual_init_obj}
\end{equation}
\noindent In addition, to stablilize training, we opt to initialize the agent's policy using BC.

\paragraph{Remark}
The root cause for virtual preferences to be unreliable here is that, our assumption $b\succ \pi$ for any $\pi$ is no longer justified.
Given preferences $\mathcal{Y}$, we expect the agent to outperform the behavior policy of $\mathcal{D}$.
Our approach reduces the problem of handling imperfect behavior policy to inferring the reliability of state-action preferences in $\mathcal{Y}_v$.

In summary, when offline trajectories are not perfect, PbAIL uses a probabilistic model for virtual preferences that can simultaneously infer the reliability of virtual preferences and learn a reward function. \Cref{alg:pb_ail} summarizes the algorithm for this case.

\begin{algorithm}[tb]
   \caption{PbAIL using off-policy RL backbone}
   \label{alg:pb_ail}
\begin{algorithmic}
   \STATE {\bfseries Input:} Offline trajectories $\mathcal{D}$, preferences $\mathcal{Y}$, number of initialization steps $K_\text{init}$.
   \STATE Initialize policy $\theta_r$, $\theta_\pi$, $\theta_\alpha$ and a replay $\mathcal{D}_\text{RL}$.
   \STATE Initialize a counter for gradient steps $k$.
   \REPEAT
        \STATE Interact with the environment and store a transition in $\mathcal{D}$.
        \STATE Sample a batch of data from $\mathcal{D}$ and $\mathcal{D}_\text{RL}$.
        \IF{$k\leq K_\text{init}$} 
            \STATE Take a minimization step for $L_\text{pref}+L_{\substack{\text{imperfect}\\\text{virtual,init}}}(\theta_r,\theta_\alpha).$
            \STATE Update the actor using the loss of BC.
        \ELSE
            \STATE Take a minimization step for $L_\text{pref}+L_{\substack{\text{imperfect}\\\text{virtual}}}(\theta_r,\theta_\alpha).$
        \ENDIF
        \STATE Take a policy improvement step using $r$ and transitions from $\mathcal{D}_\text{RL}$.
   \UNTIL{$k$ exceeds the intended number of training steps.}
\end{algorithmic}
\end{algorithm}
\section{Experiments}\label{sec:experiments}
We evaluate PbAIL from three aspects: its task performance, the individual effect of virtual preferences and preference reliability modeling, and its sensitivity against trajectory quality and preference size. 
Details regarding experiment design and implementations can be found in the appendices.

\subsection{Experiment Design}
\paragraph{Data}
We considered seven Mujoco tasks: Ant, Walker2d, Hopper, HalfCheetah (HC for short), HumanoidStandup (HS for short), Pusher, and Swimmer.
To reveal algorithms' performance in practice, we evaluated them using imperfect offline trajectories generated by multiple policies.
For each task, we used two versions of offline data: \emph{novice} and \emph{mixture}.
We first trained five SAC agents for five million steps using different random seeds and considered their final performance as the final performance for SAC.   
The novice version was generated using five model checkpoints reaching 20\% of the final performance, while the mixture version was generated by them and five checkpoints reaching 50\% of the final performance.
For preferences, we considered two sizes of $\mathcal{Y}$: $|\mathcal{Y}|=1225$ and $|\mathcal{Y}|=300$. Following prior practice~\citep{NIPS2017_d5e2c0ad}, preferences were generated for trajectory clips of length 60 using the truth rewards.



\paragraph{Alternative Methods}
We considered the method proposed by~\citet{NIPS2017_d5e2c0ad} (referred to as PbRL), which is an algorithm for online PbRL. 
We also included two other baselines, SACfD and CAIL~\citep{NEURIPS2021_670e8a43}.
SACfD is an extension to the method proposed by \citet{10.5555/3327757.3327897} that combines BC with PbRL.
CAIL is an imitation learning algorithm that encourages the ranking of offline trajectories induced by the inferred returns aligns with the given preferences.
In addition, we present the results for an ablation of the reliability estimation by using \cref{eq:virtual_obj} instead of \cref{eq:unreliable_virtual_obj}, denoted as PbAIL\textsuperscript{-}.
For reference, this paper also reports the results for SAC trained with the truth rewards (denoted as GT) and the return of offline trajectories (denoted as Data).

\paragraph{Evaluation Metrics}
The algorithms were compared for test returns after being trained for one million steps.
Returns were normalized to a 0--1 scale, such that 0 corresponds to a random policy and 1 corresponds to the final performance; a negative number indicates a method is worse than random policy. 
We report the mean and standard deviation of results for five random seeds.

\paragraph{Implementation Details}
We used SAC as the policy learner.
The reward functions of PbRL, SACfD, and PbAIL, as well as the discriminator of CAIL, were parameterized with a FFN of 64 units with the spectral normalization~\citep{miyato2018spectral}.
We also applied the weight decay to the reward functions of PbRL, SACfD, and PbAIL, and the reliability network of PbAIL.
All experiments were performed on six NVIDIA A6000 GPUs. Our code is available here\footnote{\url{https://www.dropbox.com/s/g0m3qwng4cnltmt/PbAIL.zip?dl=0}}.

\subsection{Results}

\paragraph{Task Performance} 
\Cref{tab:results_novice_all} shows the results on the novice datasets with $|\mathcal{Y}|=1225$.
First, we compare the performance of PbRL, SACfD, and PbAIL.
Except in the case of Walker2d and Swimmer, PbRL does not demonstrate competitive performance.
SACfD outperforms PbRL in Ant, Walker2d, and HC, but it does not perform well for the rest.
PbAIL surpasses PbRL in six of the seven tasks, achieving the best performance in five tasks and even outperforming GT in three tasks.
In addition, PbAIL presents significant advantages over CAIL.
These results corroborate PbAIL's efficacy for learning from offline trajectories and offline preferences.

\paragraph{Ablation Study} \Cref{tab:results_novice_all} also shows the individual effect of using virtual preferences and inferring their reliability.
The advantage of PbAIL\textsuperscript{-} over PbRL, particularly in Ant, HS, and Pusher, confirms the efficacy of using virtual preferences.
The efficacy of inferring preference reliability is supported by comparing data return (denoted as Data) and the performance of PbAIL\textsuperscript{-} accompanied with the comparison between PbAIL\textsuperscript{-} and PbAIL.
Except in Walker2d, PbAIL\textsuperscript{-} only matches the return of data, which means that information in offline preferences is not properly utilized. 
PbAIL's substantial improvement over data return confirms the necessity of modeling virtual preference reliability when dealing with imperfect offline data.

\begin{table*}[t!]
\centering
\caption{Algorithms' normalized returns for novice datasets with $|\mathcal{Y}|=1225$. We show the normalized returns of training data in  ``Data'' column and the performance obtained with the ground-truth rewards in ``GT'' column. The proposed PbAIL outperforms SACfD in five tasks and CAIL in six tasks. Notably, it even outperforms GT in three tasks. PbAIL\textsuperscript{-} is a variant of PbAIL that does not handle the non-optimality of data, and it is significantly outperformed by PbAIL. These results confirm the efficacy of learning from virtual preferences and modeling their reliability.}
\label{tab:results_novice_all}
\sisetup{table-number-alignment=center,
         detect-weight,
         separate-uncertainty = true,
         detect-inline-weight=math}
\renewcommand{\bfseries}{\fontseries{b}\selectfont}
\newrobustcmd{\tb}{\DeclareFontSeriesDefault[rm]{bf}{b}\bfseries}
\small
\begin{tabular}{@{} l *{5}{S[table-format=1.2\pm1.1,table-figures-uncertainty=1]} *{2}{S[table-format=1.2,table-figures-uncertainty=0]} @{}}
\toprule
Task        & {PbRL}               &{SACfD}      & {CAIL}   & {PbAIL\textsuperscript{-}} & {PbAIL} & {Data} & {GT} \\ \midrule
Ant & -0.35\pm0.02& 0.57\pm0.09& 0.59\pm0.09& 0.30\pm0.04& \tb 0.72\pm0.03& 0.23 & 0.66 \\
Walker2d & 0.53\pm0.17& \tb 0.61\pm0.25& 0.46\pm0.14& -0.00\pm0.00& 0.58\pm0.13& 0.28 & 0.89 \\
Hopper & 0.73\pm0.28& 0.66\pm0.25& 0.83\pm0.20& 0.31\pm0.07& \tb 1.13\pm0.01& 0.29 & 0.95 \\
HC & 0.22\pm0.23& 0.44\pm0.20& 0.74\pm0.12& 0.25\pm0.02& \tb 0.76\pm0.05& 0.22 & 0.69 \\
HS & 0.05\pm0.30& 0.05\pm0.17& 0.47\pm0.15& 0.55\pm0.16& \tb 0.64\pm0.15& 0.51 & 0.97 \\
Pusher & -4.23\pm0.72& -4.55\pm0.85& 0.22\pm0.34& 0.39\pm0.06& \tb 0.71\pm0.09& 0.37 & 0.93 \\
Swimmer & 0.19\pm0.07& 0.13\pm0.10& 0.19\pm0.10& \tb 0.30\pm0.05& 0.12\pm0.19& 0.28 & 0.64 \\
\bottomrule
\end{tabular}
\end{table*}

\paragraph{Effect of Preference Size} 
\Cref{tab:results_novice_half} shows the results for $|\mathcal{Y}|=300$ on the novice datasets.
Compared to the results presented in \cref{tab:results_novice_all}, preference-based methods (PbRL, SACfD, and PbAIL) worsen significantly, which is expected as less information is available in preferences.
Interestingly, with fewer preferences, CAIL tends to perform better—it improves in five tasks.
This is probably due to the fact that it uses preferences to update the weights of samples in $\mathcal{D}$ for imitation learning, using a max-margin loss function for preferences. 
With fewer preferences, the weights of samples can be determined more easily, so its adversarial learning process is more stable.

\begin{table*}[t!]
\centering
\caption{Performance for novice datasets with $|\mathcal{Y}|=300$.
Compared to results in \cref{tab:results_novice_all}, with less preferences, preference-based methods (PbRL, SACfD, and PbAIL) become worse, while CAIL performs better.}
\label{tab:results_novice_half}
\sisetup{table-number-alignment=center,
         detect-weight,
         separate-uncertainty = true,
		 group-digits = false,
         detect-inline-weight=math,
         table-align-text-post=false}
\renewcommand{\bfseries}{\fontseries{b}\selectfont}
\newrobustcmd{\tb}{\DeclareFontSeriesDefault[rm]{bf}{b}\bfseries}
\small
\begin{tabular}{@{} l *{4}{S[table-format=1.2\pm1.1,table-figures-uncertainty=1]} *{2}{S[table-format=1.2,table-figures-uncertainty=0]} @{}}
\toprule
Task        & {PbRL}               &{SACfD}                & {CAIL}   & {PbAIL} & {Data} & {GT} \\ \midrule
Ant & -0.37\pm0.02& -0.37\pm0.01& \tb 0.61\pm0.07& 0.56\pm0.03& 0.23 & 0.66 \\
Walker2d & 0.39\pm0.15& 0.34\pm0.16& \tb 0.49\pm0.07& 0.15\pm0.13& 0.28 & 0.89 \\
Hopper & 0.63\pm0.32& 0.38\pm0.06& 0.74\pm0.21& \tb 0.79\pm0.28& 0.29 & 0.95 \\
HC & 0.35\pm0.20& 0.12\pm0.22& \tb 0.82\pm0.08& 0.14\pm0.26& 0.22 & 0.69 \\
HS & 0.12\pm0.13& 0.18\pm0.21& 0.32\pm0.34& \tb 0.64\pm0.12& 0.51 & 0.97 \\
Pusher & -4.55\pm0.76& -4.44\pm0.75& 0.40\pm0.11& \tb 0.58\pm0.08& 0.37 & 0.93 \\
Swimmer & \tb 0.32\pm0.10& 0.26\pm0.15& 0.25\pm0.14& 0.16\pm0.27& 0.28 & 0.64 \\
\bottomrule
\end{tabular}
\end{table*}

\paragraph{Effect of Trajectory Quality} 
\Cref{tab:results_mixture_all} shows the results for the mixture datasets with $|\mathcal{Y}|=1225$. The quality of offline data increases by 0.1 to 0.2 when compared with the case in \cref{tab:results_novice_all}.
PbRL and PbAIL cannot benefit from the improved quality, but PbAIL is the only method that attains the best performance in three tasks.
SACfD improves in Ant and Walker2d but fails in HS.
CAIL can best benefit from improved quality of offline trajectories.
The results in \cref{tab:results_novice_all} and \cref{tab:results_mixture_all} imply that, PbAIL is suitable for scenarios where preferences are accessible, and imitation-based methods (SACfD and CAIL) can be better when good demonstrations can be collected.

\begin{table*}[t!]
\centering
\caption{Results on mixture datasets with $|\mathcal{Y}|=1225$. Compared to results in \cref{tab:results_novice_all}, PbRL and PbAIL cannot benefit from the improved quality of offline trajectories. SACfD improves in Ant and Walker2d, but fails in HS. CAIL has significant improvement in Ant, Walker2d, and Hopper.}
\label{tab:results_mixture_all}
\sisetup{table-number-alignment=center,
         detect-weight,
         separate-uncertainty = true,
		 group-digits = false,
         detect-inline-weight=math,
         table-align-text-post=false}
\renewcommand{\bfseries}{\fontseries{b}\selectfont}
\newrobustcmd{\tb}{\DeclareFontSeriesDefault[rm]{bf}{b}\bfseries}
\small
\begin{tabular}{@{} l *{4}{S[table-format=1.2\pm1.1,table-figures-uncertainty=1]} *{2}{S[table-format=1.2,table-figures-uncertainty=0]} @{}}
\toprule
Task        & {PbRL}               &{SACfD}                & {CAIL}   & {PbAIL} & {Data} & {GT} \\ \midrule
Ant & -0.33\pm0.05& \tb 0.74\pm0.02& 0.71\pm0.02& 0.59\pm0.03& 0.34 & 0.66 \\
Walker2d & 0.53\pm0.25& \tb 0.71\pm0.23& 0.61\pm0.02& 0.63\pm0.10& 0.38 & 0.89 \\
Hopper & 0.76\pm0.24& 0.49\pm0.16& 0.90\pm0.05& \tb 1.13\pm0.03& 0.49 & 0.95 \\
HC & 0.34\pm0.24& 0.25\pm0.28& \tb 0.74\pm0.04& 0.69\pm0.03& 0.37 & 0.69 \\
HS & 0.06\pm0.23& NaN & 0.44\pm0.17& \tb 0.62\pm0.14& 0.52 & 0.97 \\
Pusher & -5.26\pm0.45& -5.05\pm0.52& -0.10\pm0.83& \tb 0.66\pm0.04& 0.45 & 0.93 \\
Swimmer & \tb 0.38\pm0.05& 0.33\pm0.04& 0.10\pm0.19& 0.14\pm0.05& 0.40 & 0.64 \\
\bottomrule
\end{tabular}
\end{table*}

\paragraph{Reward Generalizability} Finally, let us discuss our claim on the generalizability problem.
We analyzed the generalizability of reward functions using the mixture datasets with 1225 preferences.
For each method, we took 10 model checkpoints in the early stage of policy learning.
We then rolled out trajectories using the actor of each checkpoint and inferred their returns using the corresponding reward function. 
The Kendall's rank correlation coefficient reflects how well the ranks of trajectories computed using inferred returns match those computed using ground-truth returns.
The higher it is, the better the two ranks match, which means that reward functions better generalize to the learning agent's behaviors.

\begin{figure}[t!]
     \centering
     \begin{subfigure}[t]{0.32\textwidth}
         \centering
         \includegraphics[width=\textwidth]{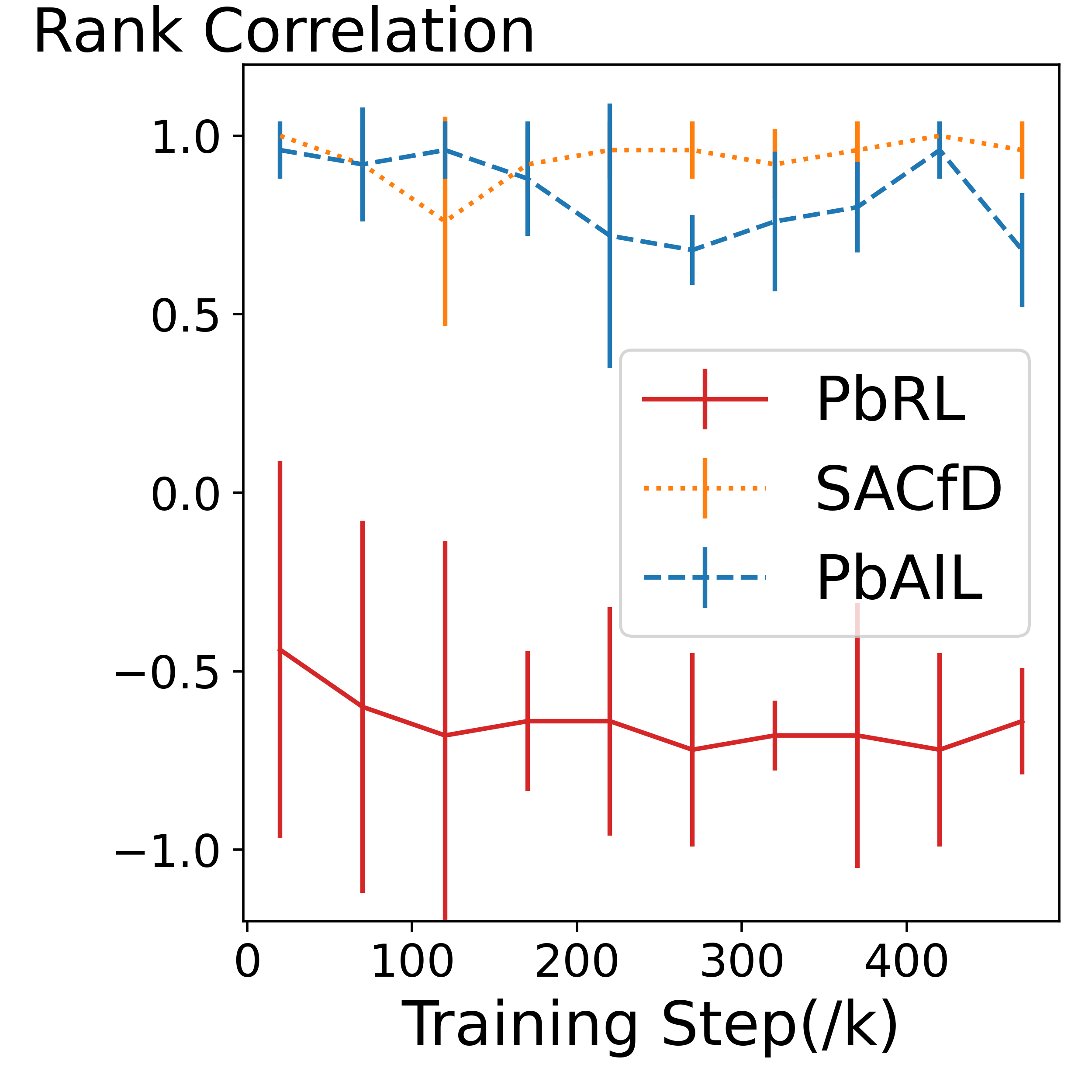}
         \caption{Ant}
     \end{subfigure}
     \begin{subfigure}[t]{0.32\textwidth}
         \centering
         \includegraphics[width=\textwidth]{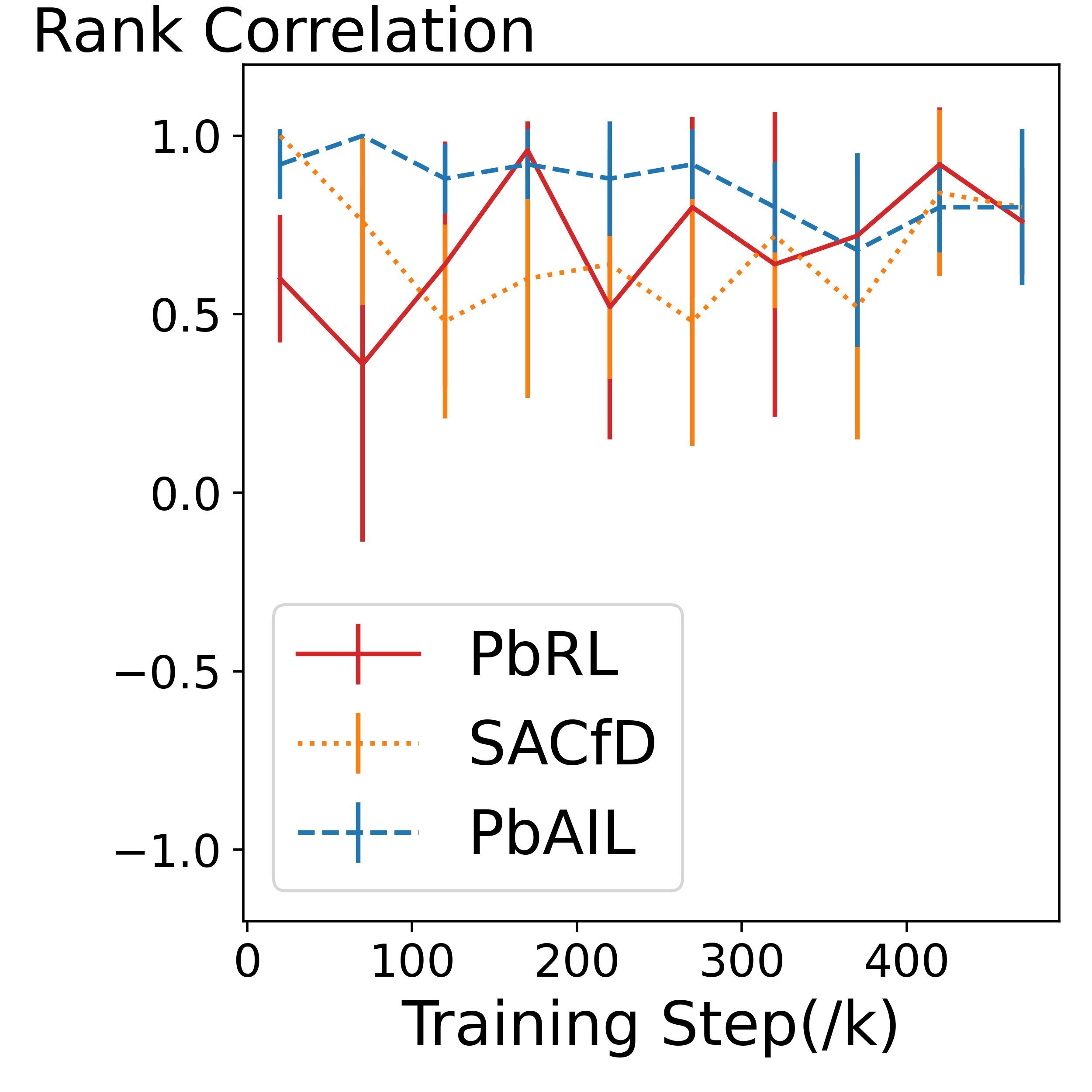}
         \caption{HC}
     \end{subfigure}
     \begin{subfigure}[t]{0.32\textwidth}
         \centering
         \includegraphics[width=\textwidth]{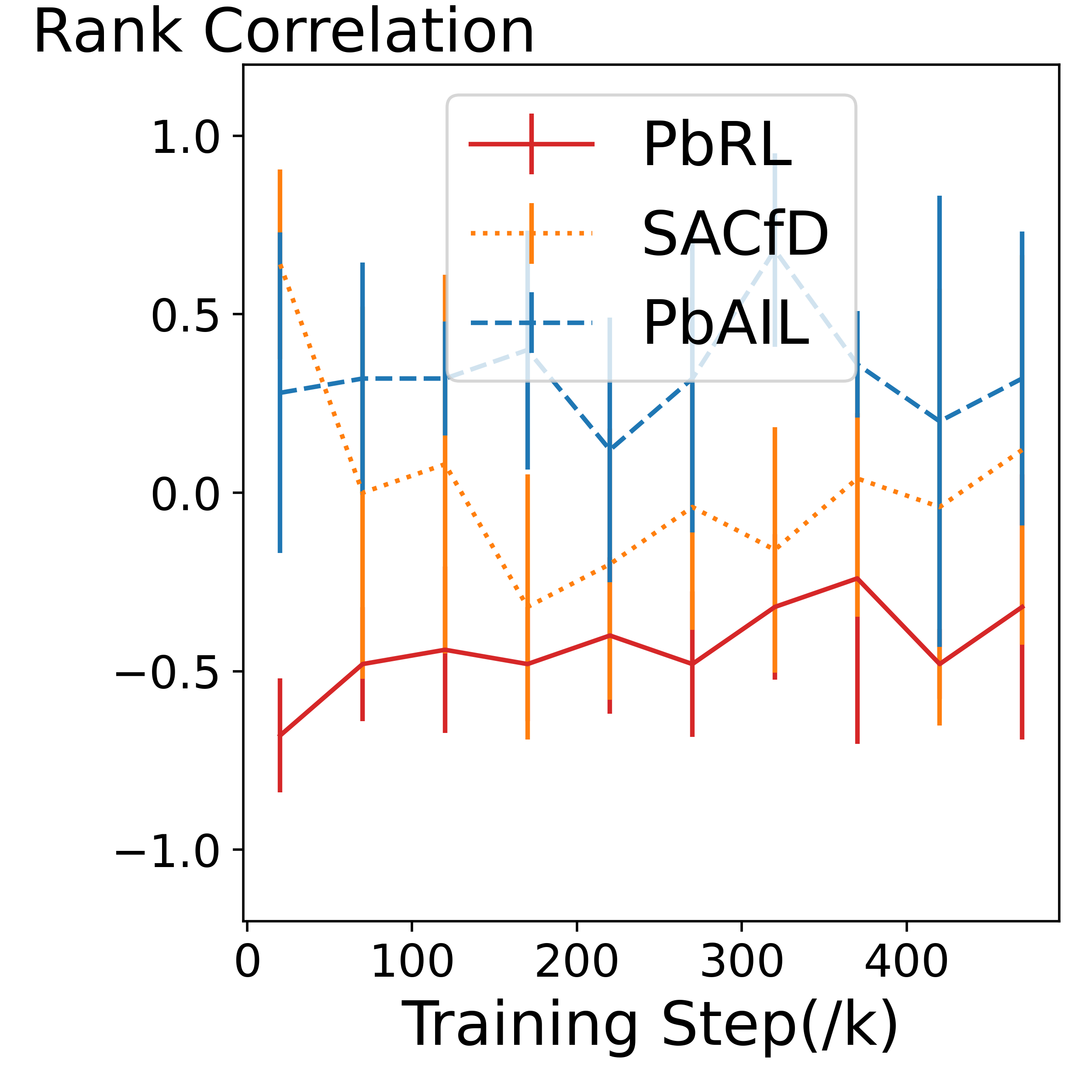}
         \caption{Pusher}
     \end{subfigure}
     \caption{The Kendall's rank correlation coefficient between the the inferred returns and the true returns of agents' trajectories during policy learning. This coefficient reflects the generalizability of reward functions to agents' behaviors. Agents were trained on the mixture datasets with $|\mathcal{Y}|=1225$. The reward function of PbRL does not generalize well for Ant and Pusher, which explains its poor performance presented in \cref{tab:results_mixture_all}. These observations support our claim for the generalizability issue.}
     \label{fig:rank_correlation}
\end{figure}

\Cref{fig:rank_correlation} shows the results for Ant, HC, and Pusher.
The reward function of PbRL fails to generalize for Ant and Pusher, which explains PbRL's poor performance for the two tasks.
Compared to PbRL, the reward function of SACfD generalizes better for Ant but fails for Pusher.
Meanwhile, for all four tasks, PbAIL's reward functions demonstrate good generalizability, which explains its performance.

In summary, our take-aways are three-folds.
\begin{itemize}
    \item The proposed PbAIL is competitive for all three cases.
    Both the use of virtual preferences and modeling their reliability work as expected.
    \item Preference-based methods tend to degenerate with fewer preferences.
    Imitation-based approaches better enjoy the improved quality of offline trajectories.
    \item Our claim for the generalizability issue of reward functions is supported by rank correlation between the inferred and true trajectory returns.
\end{itemize}
\noindent These findings confirm the efficacy of our proposals. Moreover, they shed light on how to select algorithms for complex real-world tasks. If collecting preferences is more viable than collecting good trajectories, then PbAIL is the most suitable method. Otherwise, approaches that directly imitate from trajectories are better.

\section{Conclusion}\label{sec:conclusion}
PbRL is a setting that learns a reward function from preferences, which represent human evaluation for agents' behaviors.
Recently, the use of offline preferences was proposed to make better use of human time.
In this case, the preferences are collected from certain offline data.
However, as the offline data may follow a different distribution when compared to an learning agent's performance, reward functions learned from offline preferences may fail to generalize to the agent's behaviors.
In response to this issue, the present study proposes PbAIL, a framework that overcomes this drawback by using virtual preferences generated from offline data.
PbAIL learns a reward function by jointly maximizing the likelihood of offline preferences and virtual preferences, which aligns the learned reward functions with the agent's behaviors.
Furthermore, this work extends PbAIL to imperfect offline data, thus broadening its applicability.
From experiments on continuous tasks, we verified the efficacy of using virtual preferences and handling data imperfection, and we also discussed the advantages and limitations of PbAIL.
As for future work, it would be interesting to consider extensions that do not explicitly require offline trajectories.

\bibliography{crowd_pbrl} 
\bibliographystyle{abbrvnat}

\newpage
\appendix

\section{Data}\label{app:data}
\paragraph{Offline Data}
To generate the offline data used in experiments, we first trained five SAC agents for five million steps using different random seeds and hyperparameters reported in \cref{tab:expert_hyper}.
To compute the normalized returns of algorithms, we consider the policies initialized from scratch as the random policies.
As shown in \cref{tab:behavior_performance}, their final performance matches results reported for SAC by~\citet{Haarnoja2018}.
To evaluate algorithms in realistic settings, this work considers two versions of offline data: novice and mixture. 
The novice version was generated using five model checkpoints for different random seeds that reached 20\% of the final performance.
For the mixture version, beside the five model checkpoints of the novice version, we also used five model checkpoints for different random seeds that reached 20\% of the final performance.
Both versions contain 50 trajectories, and each of the model checkpoints contributed the same amount of trajectories.

\paragraph{Preferences}
Following prior practice for PbRL~\citep{NIPS2017_d5e2c0ad}, we consider preferences between short clips of trajectories for better efficiency.
From each trajectory, we sampled one clip of length 30 in Pusher and 60 in other tasks. 
So there are 50 trajectory clips.
The case of $|\mathcal{Y}|=1225$ involves all of the paired comparisons for these 50 clips, and the case of $|\mathcal{Y}|=300$ involves paired comparisons for 25 of the 50 clips. 
The preference labels were generated using the ground-truth rewards.
In other words, for two trajectory clips $\eta_1$ and $\eta_2$, $\eta_1\succ\eta_2$ if $\sum_{x\in\eta_1}r(x)>\sum_{x\in\eta_2}r(x)$ under the ground-truth reward function $r$.

\begin{table}[h]
\caption{The final performance and performance of random policies.}
\label{tab:behavior_performance}
\centering
\small
\begin{tabular}{lccccccc}
\toprule
Task              & Ant & Walker2d & Hopper & HC & HS & Pusher & Swimmer\\ \midrule
Final Performance & 6983.98 & 4656.05 & 3097.87 & 14646.87 & 157117.92 & -20.50 & 125.57 \\
Random Policy & 989.65 & 17.08 & 46.98 & -1.44 & 42788.13 & -58.73 & 7.01 \\
\bottomrule
\end{tabular}
\end{table}

\begin{table}[h]
\caption{Hyperparameters for the behavior policies.}
\label{tab:expert_hyper}
\centering
\begin{tabular}{ll}
\toprule
Parameter &Value \\ \midrule
optimizer                                & Adam \\
learning rate for the actor              & $3\times10^{-4}$ \\
learning rate for the critic             & $3\times10^{-4}$ \\
whether to train entropy weight                   & Yes \\
initial value for entropy weight                  & 0.1 \\
learning rate for entropy weight                  & $3\times10^{-4}$ \\
\#hidden layers (all networks)                  & 2 \\
\#units per layer for the actor          & 256 \\
\#units per layer for the critic         & 256 \\
\# training steps                        & 1 \\
\# environment steps per training step   & 1\\
activation function (all networks)       & ReLU \\
interval for updating the target network & 1\\
smoothing coefficient for target updates & $5\times10^{-3}$ \\
size of replay buffer                    & $1\times10^6$ \\
discount factor                          & 0.99 \\\bottomrule
\end{tabular}
\end{table}


\section{Implementation Details}\label{app:details}
As suggested by~\citet{NEURIPS2021_7b647a7d}, we used SAC as the policy learner for all algorithms, due to its good sample efficiency.
The hyperparameters of the policy learner are the same as those for behavior policies (reported in~\Cref{tab:expert_hyper}), except for the coefficient of target updates, which is changed to 0.001 for better stability.
During training, we used the squashed Gaussian policy to enhance exploration~\citep{Haarnoja2018}; in test time we used the mode of the policy, which was deterministic.

We used the same parameterization for the reward functions of PbRL, SACfD, and PbAIL, as well as the discriminator of CAIL, so the differences in their performance could show the efficacy of our proposals.
The reward functions (discriminators) were  parameterized with a FFN of 64 units with the spectral normalization~\citep{miyato2018spectral}.
For PbRL, SACfD, and PbAIL, we also applied weight decay to their reward functions using the same but task-specific coefficients, whose values are presented in \cref{tab:l2_reward}. 
These values were selected from $\{0, 0.0025, 0.005\}$ via grid search.
The batchsize was 256 for all algorithms
We now report additional details of these methods.

\begin{table}[t]
\caption{The coefficients of weight decay for reward functions.}
\label{tab:l2_reward}
\centering
\small
\begin{tabular}{lccccccc}
\toprule
Task              & Ant & Walker2d & Hopper & HC & HS & Pusher & Swimmer\\ \midrule
Coefficient & 0.0025 & 0 & 0 & 0.0025 & 0.0025 & 0 & 0 \\
\bottomrule
\end{tabular}
\end{table}

\paragraph{Details for PbRL}
PbRL learns a reward function from preferences before policy learning and uses the learned function to infer rewards of state-action pairs generated by the policy learning.
When learning the reward function, we used the Adam optimizer with $0.0001$ as learning rate and ran optimization for 10,000 steps.

\paragraph{Details for SACfD}
SACfD is an extension to the method proposed by \citet{10.5555/3327757.3327897}.
The method was originally proposed for discrete control tasks, so we extended it for using SAC.
Compared to PbRL, it introduces a the objective of BC into the objective of the actor.
The reward function of SACfD was learned in the same way as the reward function of PbRL.
As described by \citet{10.5555/3327757.3327897}, the actor of the policy learner was pre-trained using the objective of BC and the provided offline data Before policy learning.
The pre-training took 10,000 steps.

Let $L_\text{actor}$ be the objective function for learning the actor of SAC. During policy learning, SACfD minimizes $L_\text{actor} - \mathbb{E}_{(s,a)\in\mathcal{D}}\left[\mathbb{I}(Q^\pi(s,a)>Q^\pi(s,\pi_\text{mode}(s)))\log(\pi(a|s)) \right]$, where $\mathbb{I}$ is the indicator function and $\pi_\text{mode}$ is the mode of $\pi$.
In other words, it regularizes the actor using BC only for states at which the action in the offline data has larger Q-value than the action induced by $\pi$.

\paragraph{Details for PbAIL}
As mentioned in \cref{alg:pb_ail}, PbAIL needs an initialization phase.
In all tasks we initialized PbAIL for 10,000 steps.
We applied weight decay to the reliability network in addition to its reward network.
These values were selected from $\{0, 0.0025, 0.005\}$ via grid search, and their values are reported in \cref{tab:l2_reliability}.

\begin{table}[t]
\caption{The coefficients of weight decay for the reliability network of PbAIL.}
\label{tab:l2_reliability}
\centering
\small
\begin{tabular}{lccccccc}
\toprule
Task              & Ant & Walker2d & Hopper & HC & HS & Pusher & Swimmer\\ \midrule
Coefficient & 0 & 0.005 & 0 & 0.005 & 0.005 & 0.0025 & 0.0025 \\
\bottomrule
\end{tabular}
\end{table}

\newpage

\begin{table*}[t!]
\centering
\caption{Results on real human preferences.}
\label{tab:results_mixture_all}
\sisetup{table-number-alignment=center,
         detect-weight,
         separate-uncertainty = true,
		 group-digits = false,
         detect-inline-weight=math,
         table-align-text-post=false}
\renewcommand{\bfseries}{\fontseries{b}\selectfont}
\newrobustcmd{\tb}{\DeclareFontSeriesDefault[rm]{bf}{b}\bfseries}
\small
\begin{tabular}{@{} l *{4}{S[table-format=1.2\pm1.1,table-figures-uncertainty=1]} *{2}{S[table-format=1.2,table-figures-uncertainty=0]} @{}}
\toprule
Task        & {PbRL}               &{SACfD}                & {CAIL}   & {PbAIL} & {Data} & {GT} \\ \midrule
Hopper-medium-expert & 0.57\pm0.23& 0.44\pm0.12& \tb 1.07\pm0.10& 0.65\pm0.36 & 0.67 & 0.66 \\
Hopper-medium-replay & 0.28\pm0.19& 0.28\pm0.26& 0.43\pm0.14& \tb 0.45\pm0.13 & 0.38 & 0.14 \\
Walker2d-medium-expert & 0.06\pm0.07& 0.21\pm0.05& \tb 0.90\pm0.05& 0.73\pm0.35 & 0.81 & 0.95 \\
Walker2d-medium-replay & 0.10\pm0.10& 0.08\pm0.12& \tb 0.44\pm0.15& -0.01\pm0.08 & 0.14 & 0.69 \\
\bottomrule
\end{tabular}
\end{table*}
\end{document}